\documentclass[10pt,twocolumn,letterpaper]{article}

\usepackage[pagenumbers]{cvpr} 
\usepackage{graphicx}
\usepackage{amsmath}
\usepackage{amssymb}

\usepackage{appendix}
\usepackage{booktabs}
\usepackage[pagebackref,breaklinks,colorlinks]{hyperref}

\usepackage{xcolor}
\usepackage[capitalize]{cleveref}
\crefname{section}{Sec.}{Secs.}
\Crefname{section}{Section}{Sections}
\Crefname{table}{Table}{Tables}
\crefname{table}{Tab.}{Tabs.}

\DeclareMathOperator*{\argmax}{argmax}
\usepackage{multirow}
\usepackage{pifont}

\usepackage[linesnumbered,ruled,vlined]{algorithm2e}
\SetKwInOut{KwInput}{Input}
\SetKwInOut{KwOutput}{Output}

\begin{document}

\title{Dual Prototypical Contrastive Learning for Few-shot Semantic Segmentation}
\author{Hyeongjun Kwon$^1$ \quad \quad Somi Jeong$^1$ \quad \quad Sunok Kim$^2$ \quad \quad Kwanghoon Sohn$^{1}$\\
$^1$Department of Electrical \& Electronic Engineering, Yonsei University, Seoul, Korea\\
$^2$Department of Software Engineering, Korea Aerospace University, Goyang, Korea\\
{\tt\small \{kwonjunn01, somijeong, khsohn\}@yonsei.ac.kr, sunok.kim@kau.ac.kr}}
\maketitle

\begin{abstract}
We address the problem of few-shot semantic segmentation (FSS), which aims to segment novel class objects in a target image with a few annotated samples.
Though recent advances have been made by incorporating prototype-based metric learning, existing methods still show limited performance under extreme intra-class object variations and semantically similar inter-class objects due to their poor feature representation.
To tackle this problem, we propose dual prototypical contrastive learning approach tailored to the FSS task to capture the representative semantic features effectively.
The main idea is to encourage the prototypes more discriminative by increasing inter-class distance while reducing intra-class distance in prototype feature space.
To this end, we first present a class-specific contrastive loss with a dynamic prototype dictionary that stores the class-aware prototypes during training, thus enabling the same class prototypes similar and the different class prototypes dissimilar.
Furthermore, we introduce a class-agnostic contrastive loss to enhance the generalization ability to unseen classes by compressing the feature distribution of semantic class within each episode.
We demonstrate that the proposed dual prototypical contrastive learning approach outperforms state-of-the-art FSS methods on PASCAL-$5^\text{i}$ and COCO-$20^\text{i}$ dataset.
The code is available at: \url{https://github.com/kwonjunn01/DPCL}
\end{abstract}

%%%%%%%%% BODY TEXT
\section{Introduction}
Semantic segmentation\cite{long2015fully} aims to assign the class label for every pixel in an image of its enclosing objects or regions, which has achieved remarkable progress with deep neural networks. 
Although deep networks achieve their strong performance through supervised learning, a massive amount of training data with accurate pixel-level annotations is required~\cite{everingham2010pascal,lin2014microsoft}.
When the number of labeled dataset is insufficient, it shows poor segmentation performance reducing scalability for unseen classes.
To alleviate these issues, the semantic segmentation with few-shot learning (FSL)~\cite{finn2017model,snell2017prototypical,ren2018meta} has gained an increasing attention by training models with a small amount of training dataset. 

\begin{figure}[t]
	\centering
	\includegraphics[width=0.95\columnwidth]{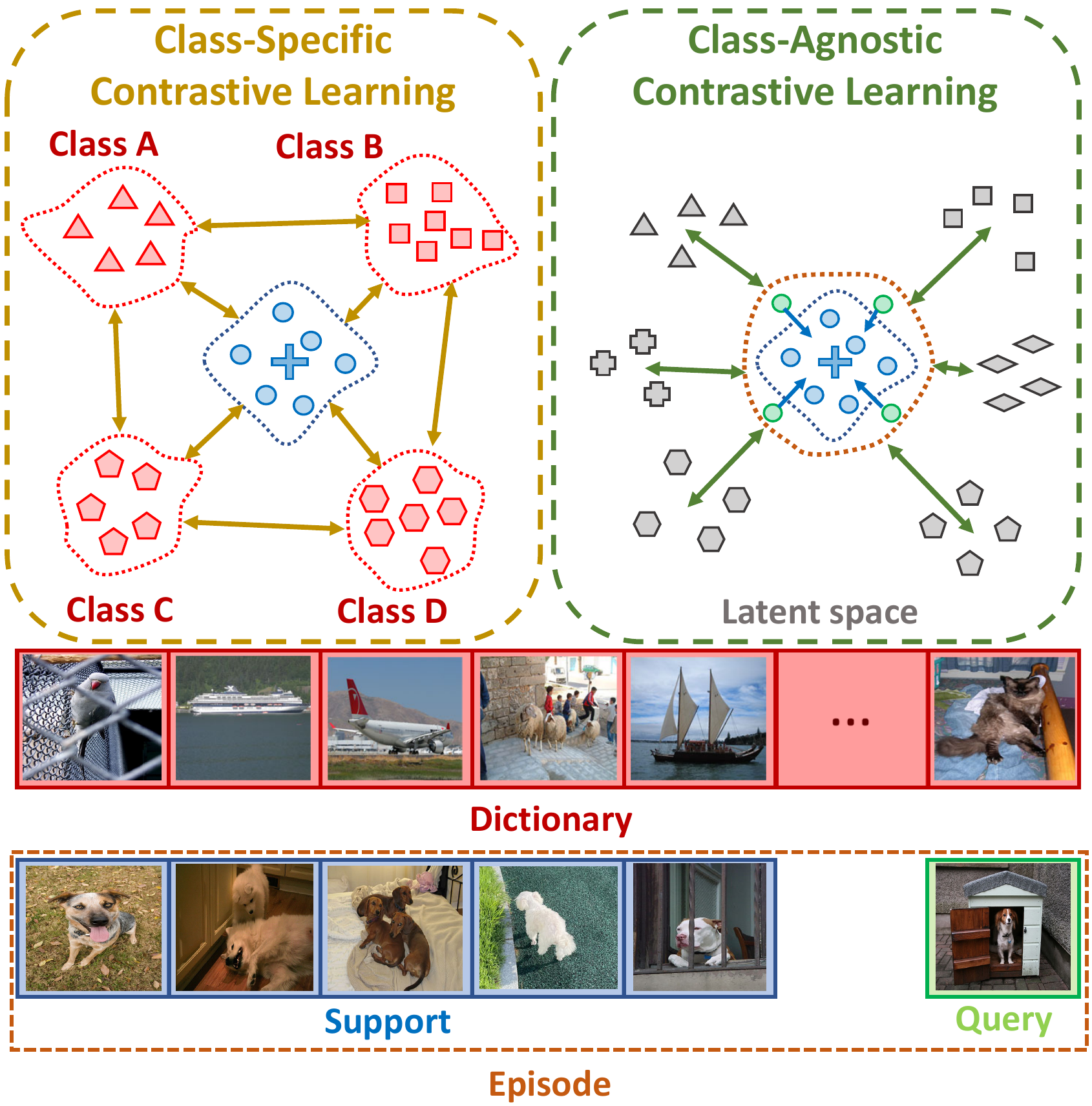}
		\caption{Key observation. Class-specific contrastive learning utilizes a dynamic prototype dictionary that stores class prototypes from past episodes to establish negative samples.
        Class-agnostic contrastive learning makes embedded features similar to its class prototypes in each episode.}
	\label{fig1}
\end{figure}

The objective of few-shot semantic segmentation (FSS) is to classify each pixel in a \emph{query} image by leveraging the information from a small number of \emph{support} images.
Recently, there have been numerous effort to solve FSS problem~\cite{wang2019panet,liu2020part,yang2020prototype} with prototype learning based metric learning~\cite{snell2017prototypical} that is widely used in the FSL with its effectiveness and simplicity.
A basic framework of the prototype-based approach is to extract the representative prototype features that encode semantic class information from the support set and to classify each pixel in the query image by comparing the distances with the class prototypes. 
Despite their improved performance, the conventional methods have a common weakness:
There is no guarantee that the prototype embeds the invariant and representative class information because they are trained by reducing only the training loss on the segmentation output without any feature-level constraint.
It may cause the problems that they can hardly handle dynamic object variations, such as large  intra-class variations of objects between the support and query images, or semantically similar inter-class objects (\eg trucks and cars) appearing in the same image.
To solve those problems, we argue that it is essential to learn good feature embedding that facilitates the inter-class feature separation and the intra-class feature compactness.

Recent advances in self-supervised representation learning~\cite{he2020momentum,chen2020simple} have reported significant improvements with contrastive loss~\cite{oord2018representation}, which is a useful objective function to learn rich and effective feature representation.
The main idea is to project the features in an embedding space that the feature distance of sample pairs belonging to the same class (\ie positive) is maximized while the distance of sample pairs from the different classes (\ie negative) is minimized.
By incorporating the contrastive loss into the FSS, it can contribute to achieve discriminative representations between the inter-class instances while maintaining invariant representations between the intra-class instances with respect to the seen classes during training.
However, the learned features are excessively discriminative against the training classes, which reduces the generalizability to unseen classes.

To address this problem, we propose \emph{Dual Prototypical Contrastive Learning} (DPCL), a new framework for FSS with representation learning to implicitly encode semantic features into the embedding space.
Our key objective is to learn discriminative representation such that a set of same class features is close to each other while the set of the different class features is far away.
To this end, we introduce two novel contrastive loss functions tailored to the FSS task for more effective intra-class and inter-class feature representation learning, as illustrated in Figure~\ref{fig1}.
First, we adopt \emph{class-specific contrastive loss} to learn the distinctive feature representation on semantic classes.
Inspired by He \etal~\cite{he2020momentum}, we leverage a dynamic prototype dictionary to store the class prototypes from each episode as a queue, which is implemented as a momentum-based moving average.
It not only facilitates the use of large-scale negatives, but also promotes the full use of limited training samples in each episode, enabling an effective prototypical contrastive learning.
Second, to generalize to the unseen class, we further suggest \emph{class-agnostic contrastive learning}, which explores feature latent spaces.
Within a single image, it focuses on minimizing the target class feature distribution and maximizing the features between the target and background features, thus increasing the adaptability to unseen classes.
With these schemes, it learns to extract the powerful features in the same classes pulled closer and the features in the different classes pushed away, thus increasing the discriminative power of learned features.
We extensively evaluate the proposed method on the standard few-shot segmentation benchmarks of PASCAL-$5^\text{i}$~\cite{shaban2017one} and COCO-$20^\text{i}$~\cite{wang2019panet}, and the experimental results demonstrate that it outperforms state-of-the-art methods without any complicated training procedure.

\section{Related work}
\subsection{Few-Shot Learning for Classification}
Few-shot learning (FSL) aims to learn transferable knowledge to new tasks with only a few data, and there has been great improvement especially in the classification task~\cite{finn2017model,ren2018meta,rusu2018meta,sun2019meta,khodadadeh2018unsupervised,nichol2018reptile}.
Generally, conventional works can be categorized into three approaches.
First, it is an optimized-based approach, including MAML~\cite{finn2017model} and Reptile~\cite{nichol2018reptile}, which aims to define specific optimization or objective functions to achieve fast learning capability.
The second is a model-based approach~\cite{santoro2016meta,munkhdalai2017meta} that facilitates the network trained quickly on new data without forgetting past data using external memory.

The last one is a metric-based approach, and its objective is to learn a good feature embedding based on an appropriate distance measurement.
Vinyals \etal~\cite{vinyals2016matching} proposed to perform for weighted nearest neighbor matching to classify unlabeled data. 
Snell \etal~\cite{snell2017prototypical} proposed a prototypical network to represent the semantic class to a single vector, in which features are used for classifying regarding distance to prototypes.
Based on the improvement in FSL, it has also been applied to other computer vision tasks, such as semantic segmentation~\cite{zhang2019canet} and object detection~\cite{kang2019few}.

\begin{figure*}[t]
		\centering
		\includegraphics[width=0.99\textwidth]{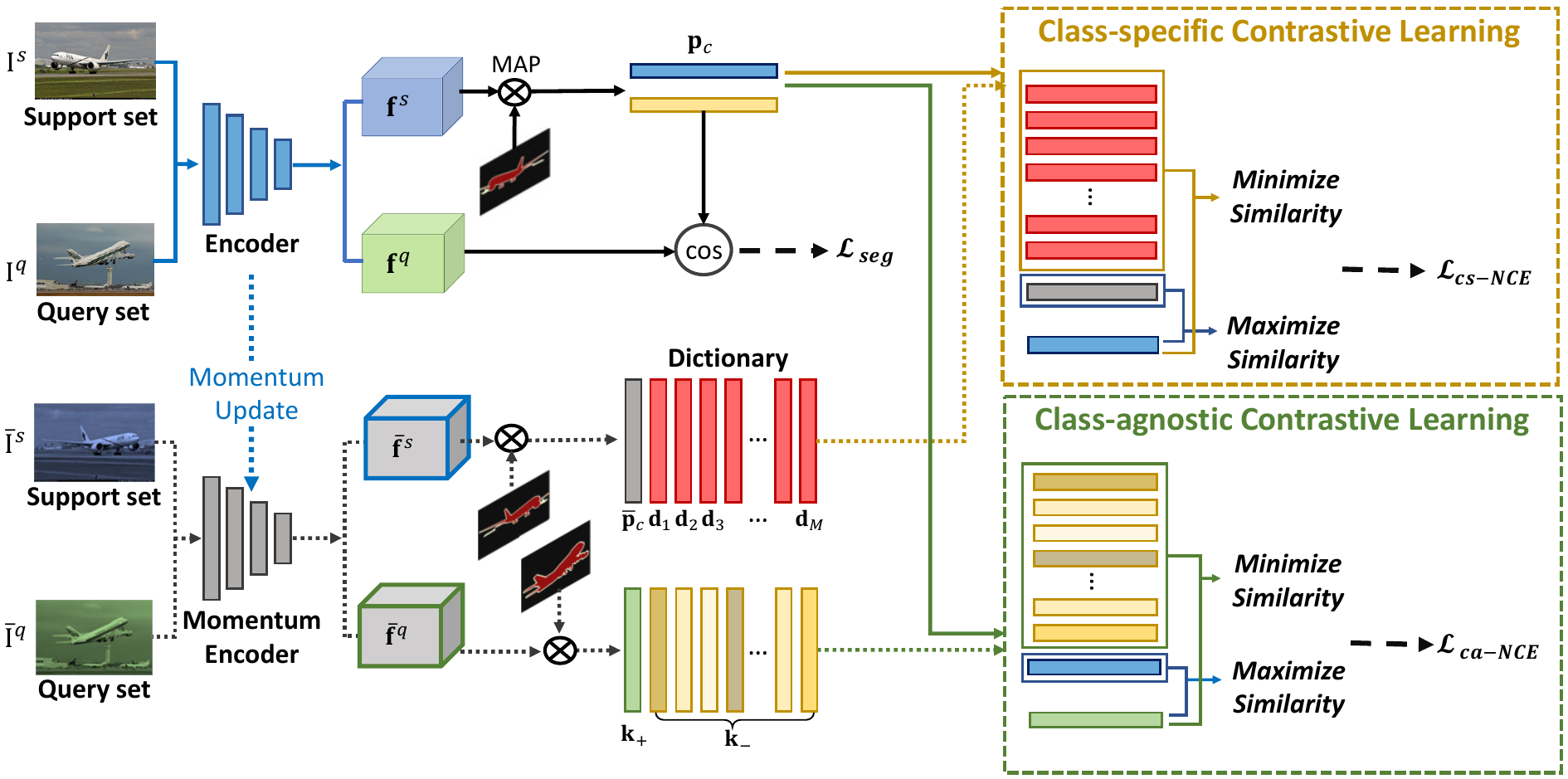}
		\caption{The framework of our proposed Dual Prototypical Contrastive Learning(\textbf{DPCL}). To learn class-specific and class-agnostic contrastive learning, we introduce contrastive learning branch, which is composed with the momentum encoder and dynamic dictionary. We design a dual contrastive objective to learn discriminative and generalized the feature representations.}
        \label{fig2}
\end{figure*}

\subsection{Few-shot Semantic Segmentation}
Few-shot semantic segmentation (FSS)~\cite{shaban2017one,rakelly2018conditional,siam2019amp,nguyen2019feature,zhang2019pyramid,liu2020part,gairola2020simpropnet,li2021adaptive,zhang2021self} has been received much attention in recent years, which benefits from a small number of annotated examples.
The objective is to predict the segmentation maps of the query image based on a few annotated support samples, which is more challenging than FSL since it performs pixel-level classification with few shots.
As a pioneer work, Shaban \etal~\cite{shaban2017one} first proposed the FSS model that generates network parameters from a conditioning branch and utilizes the obtained parameters to predict the partition map of a query set.
Zhang \etal~\cite{zhang2019canet} proposed CANet that performs the dense feature comparison between the support and query images to predict the results.
To enhance the generalization capacity to the unseen classes, recent FSS methods~\cite{wang2019panet,yang2020prototype,liu2020part,li2021adaptive,zhang2021self} have employed the prototypical network~\cite{snell2017prototypical} to solve it in non-parametric metric learning.
Specifically, it generates the class prototypes from the support images, which are used to predict the segmentation maps by conducting pixel-wise matching on query images.
These approaches mainly aim to learn better class representative prototypes.
PANet~\cite{wang2019panet} leverages a novel prototype alignment regularization to fully exploit the support knowledge, making the final predictions only by measuring the distance between the query and the prototype.
To alleviate the semantic ambiguity caused by a single prototype for a single class, PMMs~\cite{yang2020prototype} generates multiple prototypes based EM algorithm, 
and PPNet~\cite{liu2020part} exploits part-aware multiple prototypes using the clustering method.
However, the aforementioned methods cannot effectively handle the intra-class and inter-class variation problems since they do not take into account how feature representations are embedded. 

\subsection{Contrastive Learning}
Recently, the success of self-supervised learning~\cite{hjelm2018learning,oord2018representation,wu2018unsupervised,tschannen2019mutual,chen2020simple,he2020momentum} can be attributed to contrastive learning~\cite{chuang2020debiased,henaff2020data,tian2020contrastive,wang2020understanding}.
The goal of contrastive learning is to encode discriminative features effectively
by keeping samples from the same class closer and pushing samples from different classes far apart.
Recognizing that contrastive learning requires a large number of negative samples~\cite{kalantidis2020hard,chen2020simple},
He \etal~\cite{he2020momentum} proposed a momentum encoder and a memory bank to enlarge negative samples.
Chen \etal~\cite{chen2020simple} presented SimCLR that leverages various combinations of data augmentation in the current batch for sampling sufficient negative samples. 
Although many recognition tasks~\cite{khosla2020supervised,liu2021learning} benefit from this image-level contrastive learning, there is still limitation to applying this scheme directly in FSS tasks because it should be considered as pixel-level contrastive learning.
In this paper, we propose a simple yet efficient prototypical contrastive learning framework for applying the contrastive learning concept in the FSS task.

\section{Problem Formulation}
The goal of FSS is to segment novel classes objects of a set of unlabeled images (\emph{query set}) with a few annotated samples (\emph{support set}).
Following the previous FSS works~\cite{shaban2017one,wang2019panet,liu2020part}, we adopt a meta-learning strategy with episodic mechanism~\cite{vinyals2016matching}, which is effective to prevent the overfitting problem caused by lack of training data.

We briefly introduce an $N$-way $K$-shot FSS setting.
We define a training dataset $\mathcal{D}_{train}$ with base semantic classes $\mathcal{C}_{train}$, and a test dataset  $\mathcal{D}_{test}$ with unseen classes $\mathcal{C}_{test}$.
The dataset is divided into two sets, \ie a support set $\mathcal{S}$ and a query set $\mathcal{Q}$. 
Each episode $\mathcal{E}_i=\{\mathcal{S}_{i,1},\mathcal{S}_{i,2},...,\mathcal{S}_{i,K},\mathcal{Q}_i\}$ is composed of $K$ support sets from $\mathcal{S}$ and a query set from $\mathcal{Q}$.
Each support set consists of $\mathcal{S}_{i,k}=\{(\mathbf{I}_{i,k}^s, \mathbf{M}_{i,k}^s)\}$, which indicate the image and its corresponding ground-truth masks. 
Note that $\mathbf{M}_{i,k}^s$ consists of the masks of $N$ classes $\{\mathbf{M}_{i,k,1}^s, \mathbf{M}_{i,k,2}^s, \dots,\mathbf{M}_{i,k,N}^s\}$.
Then, the transferable knowledge from $\mathcal{S}_i$ is used to segment the objects in the query set $\mathcal{Q}_i=\{(\mathbf{I}_i^q, \mathbf{M}_i^q)\}$.
We exploit the ground-truth mask $\mathbf{M}_i^q$ for the supervised learning.
In testing phase, we evaluate the FSS performance using $\mathcal{D}_{test}$ across all test episodes.
Here, the training and test datasets consist of a completely different set of classes, \ie $\mathcal{C}_{train}\cap \mathcal{C}_{test}=\O$, to evaluate our effectiveness on the unseen class.

\section{Proposed Method}
We present a novel FSS framework with \emph{Dual Prototypical Contrastive Learning (DPCL)}.
The objective is to yield robust and discriminative prototypes by learning an embedding space where features of the same class are gathered and features of another class are separated.
In this section, we first describe non-parametric FSS architecture in $N$-way $K$-shot setting.
After that, we introduce the formulation of the proposed dual prototypical contrastive learning approach, \ie \emph{class-specific} and \emph{class-agnostic} contrastive learning.

\subsection{Network Architecture}
For each episode, it aims to perform the segmentation in a query set $\mathcal{Q}=\{(\mathbf{I}^q, \mathbf{M}^q)\}$ using a support set $\mathcal{S}=\{(\mathbf{I}_k^s, \mathbf{M}_k^s)\}_{k=1}^K$.\footnote{We omit the subscript $i$ for simplicity.}
Here, we denote the target unseen classes as $C=\{c_1, c_2, ... c_N\}$.
To realize this, we exploit the non-parametric FSS framework as illustrated in Figure~\ref{fig2}.

\paragraph{Feature encoder.}
Given an episode, it extracts spatial and semantic features that are used to generate the class prototype.
Since the early layers can capture low-level features (\eg edge, color, patterns), while the later layers extract more semantic features ~\cite{zeiler2014visualizing},
several approaches in object detection~\cite{lin2017feature} and semantic segmentation~\cite{long2015fully} use multiple layers to extract both low-level and high-level features.
Inspired by this, our model exploits a subset of layers along the encoder to enrich the feature representation, while the recent FSS methods use only a single feature from the encoder~\cite{wang2019panet,liu2020part}.

Concretely, the support image $\mathbf{I}_k^s$ is passed through the feature encoder $\mathcal{F}$.
Then, we apply MLP $\mathcal{H}$ to project the multi-level features in a fixed-dimensional latent and concatenate the stack of features.
The support feature is expressed as $\mathbf{f}_k^s=\{\mathcal{H}^l(\mathcal{F}^l(\mathbf{I}_k^s))\}_L$, where $(\mathcal{F}^l, \mathcal{H}^l)$ represents the $l$-th block of $(\mathcal{F}, \mathcal{H})$.
Likewise, we encode the query feature $\mathbf{f}^q$ with the set of query features $\mathbf{f}^q=\{\mathcal{H}^l(\mathcal{F}^l(\mathbf{I}^q))\}_L$.
Note that both the support and query branches leverage the shared feature encoder to enrich meaningful features for the query image as well as support images.

\paragraph{Prototype generator.}
To generate a prototype for class $c$, $\mathbf{p}_c\in\mathbb{R}^d$, a masked average pooling (MAP)~\cite{siam2019amp} is applied on $\mathbf{f}_k^s$ as:
\begin{equation}
    \mathbf{p}_c=\frac{1}{K}*\sum_{k=1}^{K}\frac{\sum_{x,y}\mathbf{f}^s_k(x,y)\cdot [\mathbf{M}_{k,c}^s(x,y) = 1]}{\sum_{x,y} [\mathbf{M}_{k,c}^s(x,y) = 1]},
    \label{eq:eq1}
\end{equation}
where $(x,y)$ is the index of the spatial position.
$\mathbf{M}_{k,c}^s$ is the binary class mask $c\in\{1,2,\dots, N+1\}$ including $N$ classes and background, where $\mathbf{M}_{k,c}^s(x,y) = 1$ indicates $(x,y)$ pixel belongs to class $c$.

\paragraph{Non-parametric learning.}
We adopt a non-parametric learning method to achieve optimal prototypes and perform segmentation well.
Using the obtained set of class prototypes $\mathcal{P}=\{\mathbf{p}_1,\mathbf{p}_2,\dots,\mathbf{p}_{N+1}\}$,
 Then we produce a probability map $\tilde{M}^{q}$ over the semantic classes, in which the probability at each pixel by measuring similarity between the query feature $\mathbf{f}^q$ and $\mathcal{P}$.
For each $\mathbf{p}_{j} \in \mathcal{P}$, we obtain the class-wise probability map $\tilde{M}^{q}_j$ as
\begin{equation}
\mathbf{\tilde{M}}^{q}_j(x,y)=\frac{\exp(-\alpha \cos(\mathbf{f}^q(x,y)),\mathbf{p}_j)}{\sum_{\mathbf{p}_j \in \mathcal{P}}\exp(-\alpha \cos(\mathbf{f}^q(x,y)),\mathbf{p}_j)},
\label{eq:eq2}
\end{equation}
where $\cos(\cdot)$ indicates the cosine distance to measure the similarity between two vectors and $\alpha$ is a temperature that enables the model to learn the best regime~\cite{oreshkin2018tadam,wang2019panet}.
Finally, the segmentation output $\mathbf{\hat{M}}^q$ is inferred as
\begin{equation}
    \mathbf{\hat{M}}^q(x,y)=\argmax_j \mathbf{\tilde{M}}^{q}_j(x,y).
\label{eq:eq3}
\end{equation}

\paragraph{Loss function.}
To train our model, we first adopt the standard segmentation loss with cross-entropy loss in the training set $\mathcal{D}_{train}$.
Following PANet~\cite{wang2019panet}, we calculate cross-entropy loss on support and query images as
\begin{equation}
\mathcal{L}_{seg} = \mathcal{L}_{CE}(\hat{\mathbf{M}}^q, \mathbf{M}^q) + \mathcal{L}_{CE}(\hat{\mathbf{M}}^s, \mathbf{M}^s),
\label{eq:lseg}
\end{equation}
where $\mathcal{L}_{CE}$ is the cross entropy loss, and $(\hat{\mathbf{M}}^q, \mathbf{M}^q)$ and $(\hat{\mathbf{M}}^s, \mathbf{M}^s)$ are the predicted and ground-truth mask of the query and support image.
Here, $\hat{\mathbf{M}}^s$ is predicted using the prototype from the query set.
It allows the information to flow in both directions between the support and the query set, thus imposing a mutual alignment.

\subsection{Dual Prototypical Contrastive Learning}
Although non-parametric FSS learning helps to learn good feature embedding, it tends to be biased and lacks the generalization capacity to unseen classes of a few training samples.
It results in low adaptability to novel classes, and especially if there are extreme object variations between the support and query images or there are objects semantically confused classes in the image. 

To this end, we propose dual prototypical contrastive learning (DPCL) as an alternative tailored to the FSS task, which aims to maximizing the inter-class feature distance while minimizing the intra-class feature variations. 
The DPCL consists of two prototype contrastive losses: \emph{class-specific contrastive loss} and \emph{class-agnostic contrastive loss}.
The class-specific contrastive loss $\mathcal{L}_{cs-NCE}$ aims to make the different class prototypes separable, thus achieving more discriminative prototypical representation for each semantic class.
We exploit a dynamic prototype dictionary that stores prototypes from past episodes to increase the similarity between the prototype and the same class prototype and reduces the similarity between different class prototypes from the dictionary.
In addition, the class-agnostic contrastive loss $\mathcal{L}_{ca-NCE}$ increases the similarity between features and their corresponding class prototype within each episode to improve the intra-class compactness in feature representation.
It helps to improve the representation ability of prototypes by aggregating ambiguous feature representations.

Inspired by MoCo~\cite{he2020momentum}, the DPCL framework contains a momentum-updated feature encoder $\mathcal{F}_m$,
whose network architecture is identical to $\mathcal{F}$.
Specifically, given support and query image $\mathbf{I}^*\in\{\mathbf{I}_k^s, \mathbf{I}^q\}$, the random data transformation $\mathcal{T}$ is applied as $\bar{\mathbf{I}}^*=\mathcal{T}(\mathbf{I}^*)$.
The augmented image $\bar{\mathbf{I}}^*$ is passed into $\mathcal{F}_m$,
and the features $\bar{\mathbf{f}}^*=\mathcal{F}_m(\bar{\mathbf{I}}^*)$ are used in the proposed class-specific and class-agnostic contrastive learning as follows.

\paragraph{Class-specific contrastive learning.}
The ideal prototype consists of invariant class representations that can be distinguished from other classes.
The previous FSS methods are trained by reducing only the training loss using a few training samples in each episode, so they can easily become overfitted based on the biased distribution formed by only a few training examples.
To alleviate this problem, we introduce the class-specific contrastive loss $\mathcal{L}_{cs-NCE}$ and the dynamic prototype dictionary to make full use of the limited training samples.
Using the stored dictionary items that consist of the prototype and its class label obtained from each training episode, we define the positive and negative prototype pairs for the current class prototype.
Then, $\mathcal{L}_{cs-NCE}$ is applied between these paired prototypes to encourage the prototype to become more discriminative.

Let us define the dynamic prototype dictionary as $\mathbf{D}=\{\mathbf{d}_u, \mathbf{l}_u\}_{u=1}^M$, where $\mathbf{d}_u$ and $\mathbf{l}_u$ indicates the stored class prototype and its class label, and $M$ is the total number of dictionary items.
For each meta-training episode, given support image $\mathbf{I}_k^s$ in the $c$-class,
we set the positive pair $(\mathbf{p}_c, \bar{\mathbf{p}}_c)$, where $\bar{\mathbf{p}}_c$ is calculated from the augmented support feature $\bar{\mathbf{f}}^s_k$.
From the stored dictionary items, we set the negative pairs $(\mathbf{p}_c, \mathbf{d}_n)$, where the class label $\mathbf{l}_u$ is not corresponding to the target $c$-class.
We then compute the cosine similarity between the paired samples
and define the following contrastive objective function as
\begin{equation}
    \mathcal{L}_{cs-NCE} = -\log \frac{\exp(\cos(\mathbf{p}_c, \bar{\mathbf{p}}_c)/\tau)}{\sum_{u=1}^{U} \exp (\cos(\mathbf{p}_c, \mathbf{d}_u)/\tau)},
    \label{eq:localloss}
\end{equation}
where $\tau$ is a temperature and $U$ means the number of the negative pairs.
After calculating the class-specific contrastive loss, the $c$-class prototype and its class label is put into the dictionary and the oldest item is dequeued. 

\paragraph{Class-agnostic contrastive learning.}
To attain the generalization ability to unseen classes, the feature should be embedded to strengthen the semantic characteristics.
However, since $\mathcal{L}_{cs-NCE}$ focuses only on making the prototype composed of the first-order statistic of a feature set distinctive, it is vulnerable to noise such as large intra-class variations of objects in the same class or semantically similar objects of different classes.
Therefore, we introduce the class-agnostic contrastive loss $\mathcal{L}_{ca-NCE}$ to aggregate pixel-wise class features to their class prototypes, leading to the compact feature distribution for each class.

Concretely, given $\mathbf{p}_c$, we build the positive and negative samples using the encoded query feature $\bar{\mathbf{f}}^q$ from the transformed query image $\bar{\mathbf{I}}^q$.
To enhance the intra-class feature compactness by aggregating the features belonging to the same class, we set the positive sample $\mathbf{k}_+$ generated by MAP of $\bar{\mathbf{f}}^q$ in the $c$-class.
Since the robust features can be learned by a rich set of negative samples~\cite{he2020momentum}, we additionally adopt a new strategy to augment the negative samples.
Instead of applying MAP on the entire negative pixels, we randomly select the negative pixels and average them to build a set of negative samples $\mathbf{k}_-=\{\mathbf{k}_1, \mathbf{k}_2, \dots, \mathbf{k}_L\}$.
Similar to $\mathcal{L}_{cs-NCE}$, we apply the contrastive loss between $\mathbf{p}_c$ and positive and negative samples as
\begin{equation}
    \mathcal{L}_{ca-NCE} = -\log \frac{\exp(\cos(\mathbf{p}_c, \mathbf{k}_+)}{\sum_{v=1}^{V} \exp ((\cos(\mathbf{p}_c, \mathbf{k}_{v})/\tau)},
    \label{eq:caNCE}
\end{equation}
where $V$ is the number of negative samples.

\begin{algorithm*}[t!]
  \caption{Training step}
  \label{alg1}
  \SetAlgoLined
  \KwInput{$\mathcal{D}_{train}\in\{\mathcal{S}, \mathcal{Q}\}$ where support set $\mathcal{S}=\{(\mathbf{I}^s, \mathbf{M}^s)\}$, query set $\mathcal{Q}=\{(\mathbf{I}^q, \mathbf{M}^q)\}$; \\
  dynamic prototype dictionary $\mathbf{D} = \left\{\mathbf{d}_u,\mathbf{l}_u\right\}^{M}_{u=1}$; momentum encoder $\mathcal{F}_m$ with parameters $\theta_m$.}
  \KwOutput{FSS model $\mathcal{F}$ with parameters $\theta$.}
  \BlankLine
  \Begin{
     Initializing parameters $\theta$ and setting $\theta_m \leftarrow \theta$\\
     Initializing dictionary items $\mathbf{d}_u, \mathbf{l}_u$ with random variable\\
     \For {$\{\mathcal{S}, \mathcal{Q}\} \leftarrow \mathcal{D}_{train}$ \textbf{to} maximum iteration}
     { $\mathcal{L}_{seg} = FSS(\mathcal{S}, \mathcal{Q}; \theta)$\\
      $\mathcal{L}_{cs-{NCE}}, \mathbf{D} = CSCL(\mathcal{S}, \mathbf{D}; \theta, \theta_m)$\\
      $\mathcal{L}_{ca-{NCE}} = CACL(\mathcal{S}, \mathcal{Q}; \theta, \theta_m)$\\
      $\mathcal{L} \leftarrow \mathcal{L}_{seg} + \lambda_1\mathcal{L}_{cs-NCE} + \lambda_2\mathcal{L}_{ca-NCE}$\\
      Update $\theta$ with SGD by minimizing $\mathcal{L}$\\
      Update $\theta_m$ with momentum-based moving average $~~\theta_{m} \leftarrow m\theta_m + (1-m)\theta$\\}
     }
\end{algorithm*}

\paragraph{Momentum update.}
It is difficult to update the momentum encoder $\mathcal{F}_m$ by back-propagation as the dynamic prototype dictionary contains considerable prototypes to provide a large number of negative samples.
To solve this problem, we adopt the momentum update approach, which is updated with an exponential moving average of the encoder weights.
Formally, we define the parameters of encoder as $\theta$ and those of momentum encoder as $\theta_{m}$.
We update $\theta_m$ as follows:
\begin{equation}
    \theta_{m} \leftarrow m\theta_m + (1-m)\theta,
    \label{eq:Moup}
\end{equation}
where $m \in [0,1)$ means a momentum coefficient.
This makes the momentum encoder updated slowly during training, ensuring consistent feature representations from the augmented images.
As a result, it encourages the encoder to learn optimal feature embedding.

\vspace{5pt}
To sum up, the total loss for training our model is
\begin{equation}
\mathcal{L} = \mathcal{L}_{seg} + \lambda_1\mathcal{L}_{cs-NCE} + \lambda_2\mathcal{L}_{ca-NCE},
    \label{eq:total}
\end{equation}
where $\lambda_1$ and $\lambda_2$ control the importance of each term.

\begin{table*}[h]
\centering
\resizebox{\textwidth}{!}{% 
\begin{tabular}{lcccccccccccc} 
\toprule
\multirow{2}{*}{Methods}& \multirow{2}{*}{Dec} & \multirow{2}{*}{Backbone}  & \multicolumn{5}{c}{1-shot}           & \multicolumn{5}{c}{5-shot}             \\ 
\cmidrule(lr){4-8}\cmidrule(lr){9-13}
                         &&                            & fold-1  & fold-2  & fold-3  & fold-4  & \textbf{Mean} & fold-1  & fold-2  & fold-3  & fold-4  & \textbf{Mean}   \\ 
\hline\midrule
FWB~\cite{nguyen2019feature}          &\checkmark& \multirow{1}{*}{VGG-16}    & 47.04 & 59.64 & 52.61 & 48.27 & 51.90 & 50.87 & 62.86 & 56.48 & 50.09 & 55.08  \\ 
\midrule
CANet~\cite{zhang2019canet}            &\checkmark& \multirow{7}{*}{ResNet-50} & 52.50 & 65.90 & 51.30 & 51.90 & 55.40 & 55.50 & 67.80 & 51.90 & 53.20 & 57.10  \\
PGNet~\cite{zhang2019pyramid}          &\checkmark&                            & 56.00 & 66.90 & 50.60 & 50.40 & 56.00 & 57.70 & 68.70 & 52.90 & 54.60 & 58.50  \\
RPMMs~\cite{yang2020prototype}         &\checkmark&                            & 55.15 & 66.9  & 52.61 & 50.68 & 56.34 & 56.28 & 67.34 & 54.52 & 51.00 & 57.30  \\
SimPropNet~\cite{gairola2020simpropnet}&\checkmark&                            & 54.82 & 67.33 & 54.52 & 52.02 & 57.19 & 57.20 & 68.50 & 58.40 & 56.05 & 60.04  \\
PFENet~\cite{tian2020prior}            &\checkmark&                            & 61.70 & 69.50 & 55.40 & 56.30 & 60.80 & 63.10 & 70.70 & 55.80 & 57.90 & 61.90  \\
ASGNet~\cite{li2021adaptive}           &\checkmark&                            & 58.84 & 67.86 & \textbf{56.79} & 53.66 & 59.29 & 63.66 & 70.55 & \textbf{64.17} & 57.38 & 63.94  \\
SCL~\cite{zhang2021self}               &\checkmark&                            & \textbf{63.00} & \textbf{70.00} & 56.50  & \textbf{57.7}  & \textbf{61.8}  & \textbf{64.50}  & \textbf{70.90}  & 57.30  & \textbf{58.70}  & \textbf{62.90}  \\
\midrule[1pt]
OSLSM~\cite{shaban2017one}             &\ding{55}& \multirow{4}{*}{VGG-16}    & 33.60 & 55.30 & 40.90 & 33.50 & 40.80 & 35.90 & 58.10 & 42.10 & 39.10 & 43.95  \\
co-FCN~\cite{rakelly2018conditional}   &\ding{55}&                            & 36.70 & 50.60 & 44.90 & 32.40 & 41.10 & 37.50 & 50.00 & 44.10 & 33.90 & 41.40  \\
AMP~\cite{siam2019amp}            &\ding{55}&                            & 41.90 & 50.20 & 46.70 & 34.40 & 43.40 & 40.30 & 55.30 & 49.90 & 40.10 & 46.40  \\
PANet~\cite{wang2019panet}             &\ding{55}&                            & 42.30 & 58.00 & 51.10 & 41.20 & 48.10 & 51.80 & 64.60 & 59.80 & 46.50 & 55.70  \\
\midrule
PANet~\cite{wang2019panet}          &\ding{55}&\multirow{3}{*}{ResNet-50}  & 44.03 & 57.52 & 50.84 & 44.03 & 49.10 & 55.31 & 67.22 & 61.28 & 53.21 & 59.26  \\
PPNet~\cite{liu2020part}          &\ding{55}&                            & 47.83 & 58.75 & \textbf{53.80} & 45.63 & 51.50 & 58.39 & 67.83 & \textbf{64.88} & 56.73 & 61.96  \\
\textbf{Ours (DPCL)}                            &\ding{55}&                            & \textbf{52.46}  & \textbf{61.56} & 52.31 & \textbf{47.18} & \textbf{53.39} & \textbf{60.92} & \textbf{71.79} & 62.60 & \textbf{57.12} & \textbf{63.11}   \\
\bottomrule
\end{tabular}}
\caption{Comparison with state-of-the-art methods on PASCAL-$5^\text{i}$. We divide the conventional methods based on `with' and `without' decoder network. 
The best performances in each type are highlighted in bold.}
\label{tab:voc}
\end{table*}

\section{Experiments}
\subsection{Experimental Settings}
\paragraph{Dataset.} 
We use two standard FSS benchmarks, PASCAL-$5^\text{i}$~\cite{shaban2017one} and COCO-$20^\text{i}$~\cite{wang2019panet}.
The PASCAL-$5^\text{i}$~\cite{shaban2017one} dataset contains 20 classes, which is constructed from PASCAL VOC 2012 dataset~\cite{everingham2010pascal} with SBD augmentation~\cite{hariharan2011semantic}.
The COCO-$20^\text{i}$~\cite{wang2019panet} consists of 80 classes created from MS COCO dataset~\cite{lin2014microsoft}.
In the experiment, we use the same data split protocols as PANet~\cite{wang2019panet}.
For PASCAL-$5^\text{i}$ and COCO-$20^\text{i}$\footnote{We do not experiment with COCO-$20^\text{i}$ data split in FWB.}, the selected 15 and 60 classes as training data, and the remaining 5 and 20 classes as test data, respectively.
On both datasets, we apply cross-validation schemes by evenly splitting the data in four folds. 
That is, three folds are used in the training stage, and the remaining fold is used in the test stage.

\paragraph{Implementation details.}
We implement the proposed method based on PANet~\cite{wang2019panet}\footnote{https://github.com/kaixin96/PANet}, which we treat as the baseline method.
The feature encoder is composed of ResNet-50~\cite{he2016deep} and three one-layer MLPs to make the intermediate features extracted from each block of ResNet-50 into 512 dimensions.
The momentum encoder has the same network architecture as the feature encoder.
We trained the proposed model on a single NVIDIA TITAN RTX GPU.
We set the batch size 2, and use SGD optimizer with the initial learning rate 1e-3, where the momentum and weight decay are 0.9 and 5e-4, respectively.

For the FSS non-parameter learning, we set $\alpha$ as 20, following the PANet~\cite{wang2019panet}.
For the dual prototypical contrastive learning, we empirically set $\lambda_{1}$ and $\lambda_{2}$ to 0.02 and 0.015, and $\tau$ and $m$ as 0.05 and 0.999.
The total number of the items in the dynamic dictionary ${D}$ is set to $M$=8,192, and the number of negative samples in $\mathcal{L}_{cs-NCE}$ and $\mathcal{L}_{ca-NCE}$ are set to $J$=7,000 and $V$=1,000.

The images are cropped to 473 $\times$ 473 resolution.
As the random data transformation to augment the input image, we follow MoCo v2~\cite{chen2020improved}, including random crop, random color jittering, random horizontal flip, random grayscale conversion, and random blur augmentation.

\paragraph{Evaluation metric and comparison methods.}
For evaluation, we measure a mIoU metric that averages Intersection-over-Union (IoU) of all
the classes to compare the accuracy and generalizability.
To demonstrate the effectiveness of our method, we extensively compare it with other methods.
We divide these methods into two types based on \emph{`with'} and \emph{`without'} the decoder network.
Note the decoder network takes the matching probability and predicts the segmentation mask from the enhanced probability, which results in the improved performance despite the high complexity.
We categorize the conventional methods into two types as:
\begin{itemize}
    \item \emph{w/ decoder}: FWB~\cite{nguyen2019feature}, CANet~\cite{zhang2019canet}, PGNet~\cite{zhang2019pyramid}, RPMMs~\cite{yang2020prototype}, SimPropNet~\cite{gairola2020simpropnet}, PFENet~\cite{tian2020prior}, ASGNet~\cite{li2021adaptive}, SCL~\cite{zhang2021self}
    \item \emph{w/o decoder}: OSLSM~\cite{shaban2017one}, co-FCN~\cite{rakelly2018conditional}, AMP~\cite{siam2019amp}, PANet~\cite{wang2019panet}, PPNet~\cite{liu2020part}
\end{itemize}

\begin{table*}[t]
\centering
\resizebox{1\linewidth}{!}{
\begin{tabular}{lcccccccccccc}
\toprule
\multirow{2}{*}{Methods} & \multirow{2}{*}{Dec}& \multirow{2}{*}{Backbone}  & \multicolumn{5}{c}{1-shot}      & \multicolumn{5}{c}{5-shot}        \\ 
\cmidrule(lr){4-8}\cmidrule(lr){9-13}
                         &         &                   & fold-1  & fold-2  & fold-3  & fold-4  & \textbf{Mean} & fold-1  & fold-2  & fold-3  & fold-4  & \textbf{Mean} \\ 
\hline \midrule
PANet~\cite{wang2019panet}  &\ding{55}&  VGG-16                     & 28.7 & 21.2 & 19.1 & 14.8 & 20.9 & 39.4 & 28.3 & 28.2 & 22.7 & 29.7  \\
\midrule
PANet~\cite{wang2019panet}       &\ding{55}&           \multirow{3}{*}{ResNet-50} & 31.5 & 22.6 & 21.5 & 16.2 & 22.9 & 33.0 & 45.9 & 29.2 & 30.6 & 33.8  \\
PPNet~\cite{liu2020part}&\ding{55}&                                           & 34.5 & 25.4 & 24.3 & 18.6 & 25.7 & 48.3 & 30.1 & \textbf{36.7} & 30.2 & 36.2  \\
\textbf{Ours (DPCL)}                    &\ding{55}&           & \textbf{37.4} & \textbf{41.4} & \textbf{29.8} & \textbf{26.5} & \textbf{33.4} & \textbf{50.9} & \textbf{43.1} & 34.3 & \textbf{31.3} & \textbf{39.9} \\
\bottomrule
\end{tabular}}
\caption{Comparison with state-of-the-art methods w/o decoder on COCO-$20^\text{i}$.
The best performances are highlighted in bold.}
\label{tab:coco}
\end{table*}

\begin{figure}[t!]
     \centering
     \begin{subfigure}[b]{0.19\linewidth}
         \centering
         \includegraphics[width=\textwidth]{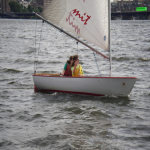}
     \end{subfigure}
     \hfill
     \begin{subfigure}[b]{0.19\linewidth}
         \centering
         \includegraphics[width=\textwidth]{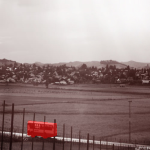}
     \end{subfigure}
     \hfill
     \begin{subfigure}[b]{0.19\linewidth}
         \centering
         \includegraphics[width=\textwidth]{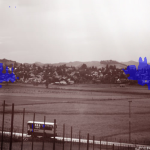}
     \end{subfigure}
     \hfill
     \begin{subfigure}[b]{0.19\linewidth}
         \centering
         \includegraphics[width=\textwidth]{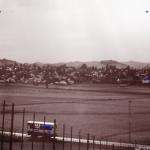}
     \end{subfigure}
     \hfill
     \begin{subfigure}[b]{0.19\linewidth}
         \centering
         \includegraphics[width=\textwidth]{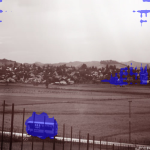}
     \end{subfigure}
    
    \begin{subfigure}[b]{0.19\linewidth}
         \centering
         \includegraphics[width=\textwidth]{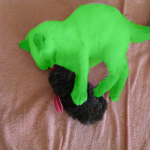}
     \end{subfigure}
     \hfill
     \begin{subfigure}[b]{0.19\linewidth}
         \centering
         \includegraphics[width=\textwidth]{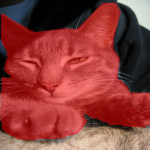}
     \end{subfigure}
     \hfill
     \begin{subfigure}[b]{0.19\linewidth}
         \centering
         \includegraphics[width=\textwidth]{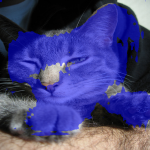}
     \end{subfigure}
     \hfill
     \begin{subfigure}[b]{0.19\linewidth}
         \centering
         \includegraphics[width=\textwidth]{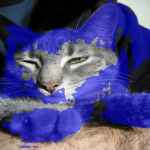}
     \end{subfigure}
     \hfill
     \begin{subfigure}[b]{0.19\linewidth}
         \centering
         \includegraphics[width=\textwidth]{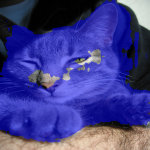}
     \end{subfigure}
    
     \begin{subfigure}[b]{0.19\linewidth}
         \centering
         \includegraphics[width=\textwidth]{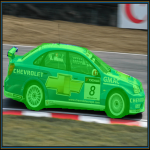}
     \end{subfigure}
     \hfill
     \begin{subfigure}[b]{0.19\linewidth}
         \centering
         \includegraphics[width=\textwidth]{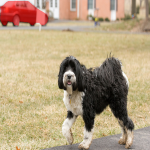}
     \end{subfigure}
     \hfill
     \begin{subfigure}[b]{0.19\linewidth}
         \centering
         \includegraphics[width=\textwidth]{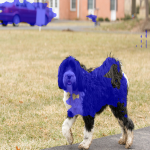}
     \end{subfigure}
     \hfill
     \begin{subfigure}[b]{0.19\linewidth}
         \centering
         \includegraphics[width=\textwidth]{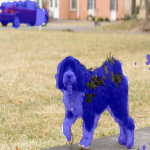}
     \end{subfigure}
     \hfill
     \begin{subfigure}[b]{0.19\linewidth}
         \centering
         \includegraphics[width=\textwidth]{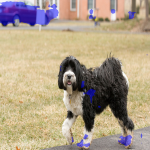}
     \end{subfigure}
     
     \begin{subfigure}[b]{0.19\linewidth}
         \centering
         \includegraphics[width=\textwidth]{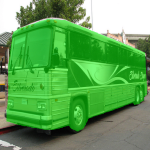}
         \caption{Support}
     \end{subfigure}
     \hfill
     \begin{subfigure}[b]{0.19\linewidth}
         \centering
         \includegraphics[width=\textwidth]{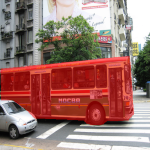}
         \caption{Query}
     \end{subfigure}
     \hfill
     \begin{subfigure}[b]{0.19\linewidth}
         \centering
         \includegraphics[width=\textwidth]{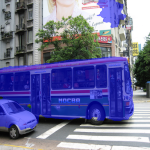}
         \caption{PANet}
     \end{subfigure}
     \hfill
     \begin{subfigure}[b]{0.19\linewidth}
         \centering
         \includegraphics[width=\textwidth]{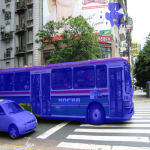}
         \caption{PPNet}
     \end{subfigure}
     \hfill
     \begin{subfigure}[b]{0.19\linewidth}
         \centering
         \includegraphics[width=\textwidth]{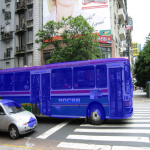}
         \caption{Ours}
     \end{subfigure}
        \caption{Results in 1-way 1-shot setting on PASCAL-$\mathbf{5}^i$.}
        \label{fig:qr}
\end{figure}

\subsection{Comparisons to State-of-the-art}
We conduct evaluations on PASCAL-$5^\text{i}$ and COCO-$20^\text{i}$ datasets in comparison with recent methods.
For a fair comparison, we compare them depending on with and without decoder network.

\paragraph{PASCAL-$5^\text{i}$.}
In Table~\ref{tab:voc}, we show the performance comparison for 1-shot and 5-shot settings.
As can be seen, our method surpasses the previous methods without decoder by large margins.
Especially, in the 5-shot setting, compared to the baseline method, PANet~\cite{wang2019panet}, our method gains mIoU increases range from about 4\% without any use of extra unlabeled training data.
These results confirm that our DPCL can learn the optimal embedding that contains invariant class representative features.
Moreover, compared to the methods with the decoder network, our method shows comparable results, especially in the 5-shot setting.
In Figure~\ref{fig:qr}, we show the qualitative results.
We observe that our method can generate precise segmentation results including ambiguous and difficult areas compared to other methods.

\paragraph{COCO-$20^\text{i}$.}
Our method outperforms the conventional methods by a large margin of 10.5\% and 6.1\% on 1-shot and 5-shot setting, respectively, as shown in Table~\ref{tab:coco}.
Although the COCO-$20^\text{i}$ has more object classes and challenging samples, our method shows robust results compared to other methods thanks to effective contrast learning.
This proves that our approach shows superior performance than other methods regardless of the difficulty of the dataset.

\subsection{Ablation Study}
We examine the impact of class-specific and class-agnostic contrastive learning on the prototypes.
We also study how concatenating the feature layer prevents overfitting and helps to extract informative features.
We conduct all ablation studies in 1-way 1-shot setting on PASCAL-$5^\text{i}$ dataset.

\begin{table}[t]
\centering
\resizebox{\linewidth}{!}{% 
\begin{tabular}{cc|ccccc}
\toprule
~CSCL~ & ~CACL~ & \multicolumn{1}{l}{fold-1}   & \multicolumn{1}{l}{fold-2} & \multicolumn{1}{l}{fold-3} & \multicolumn{1}{l}{fold-4} & \multicolumn{1}{l}{\textbf{Mean}} \\ \hline\midrule
- & -                   & 44.64          & 57.95          & 51.02          & 45.03          & 49.65   \\
- & \checkmark          & 46.54          & 59.90          & 51.11          & 45.15          & 50.68   \\
\checkmark & -          & 47.30          & 58.05          & 50.90          & 46.60          & 50.71   \\
\checkmark & \checkmark & \textbf{52.46} & \textbf{61.56} & \textbf{52.31} & \textbf{47.18} & \textbf{53.39}        \\
\bottomrule
\end{tabular}}
\caption{Ablation study about class-specific (CSCL) and class-agnostic contrastive learning (CACL).}
\label{tab:cl}
\end{table}
\begin{table}[t]
\centering
\resizebox{\linewidth}{!}{
\begin{tabular}{ccc|ccccc}
\toprule
~$l2$~ & ~$l3$~ & ~$l4$~ & \multicolumn{1}{l}{fold-1}   & \multicolumn{1}{l}{fold-2} & \multicolumn{1}{l}{fold-3} & \multicolumn{1}{l}{fold-4} & \multicolumn{1}{l}{\textbf{Mean}} \\ \hline\midrule
- &- & \checkmark                        & 49.11            & 61.90                   & 50.07           & 46.42                   & 51.88                       \\
- & \checkmark& \checkmark               & 50.33            & \textbf{62.16}          & 51.84           & 45.33                   & 52.44                       \\
\checkmark &\checkmark & \checkmark      & \textbf{52.46}   & 61.56                   & \textbf{52.31}  & \textbf{47.18}          & \textbf{53.39}                      \\ \bottomrule
\end{tabular}}
\caption{Ablation study of the different fusion of layers.}
\label{tab:mlf}
\end{table}

\paragraph{Effectiveness of contrastive learning.}
In Table~\ref{tab:cl}, it shows that applying both losses has the best performance. 
Although each element alone improved the performance, when applied at the same time, the performance improved by 3.7\% compared to the baseline method. 
These results prove that class-specific and class-agnostic contrastive learning is mutually beneficial in learning more discriminative and generalized feature representations of each semantic class.

\begin{figure}[t!]
	\centering
	\begin{subfigure}[b]{0.49\linewidth}
         \centering
         \includegraphics[width=\textwidth]{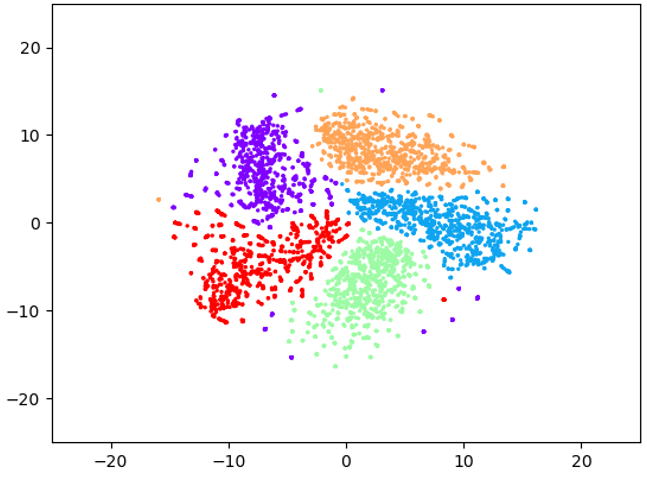}
         \caption{Baseline}
     \end{subfigure}
     \hfill
     \begin{subfigure}[b]{0.49\linewidth}
         \centering
         \includegraphics[width=\textwidth]{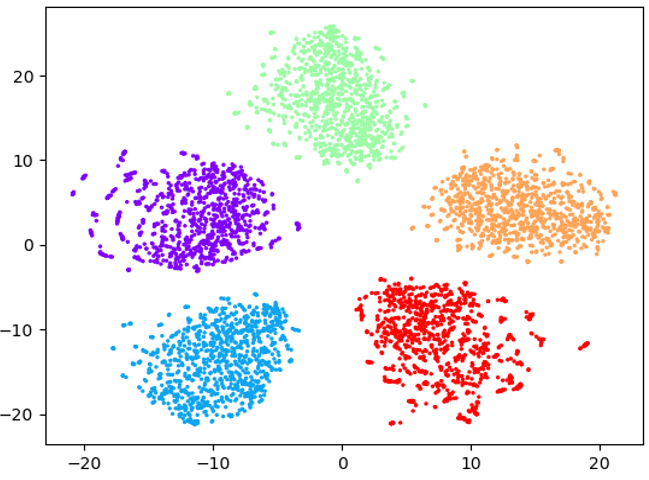}
         \caption{DPCL}
     \end{subfigure}
		\caption{The t-SNE visualization of novel class prototypes from the baseline and the proposed DPCL.}
	\label{fig:tsne}
\end{figure}

To investigate its effect on the prototype, we additionally visualize the prototype distribution of unseen classes in the test set.
We project the obtained prototypes into 2D space using t-SNE~\cite{van2008visualizing}.
As shown in Figure~\ref{fig:tsne}, the learned prototype embeddings by DPCL become more compact and well separated compared to the baseline.
This supports that the proposed DPCL can produce better segmentation performance for the unseen classes.

\paragraph{Effectiveness of multi-layer feature fusion.}
To explore the effect of feature fusion, we conduct experiments with different combinations of feature in ResNet-50~\cite{he2016deep}.
As shown in Table~\ref{tab:mlf}, using multiple features shows better performance than using a single feature since it can take advantage of abundant feature representation provided by multiple layers.
We achieve the best result using the combination of features from layer 2, layer3, and layer4 because it allows incorporating semantic cues from low-level to high-level features, thus improving the prediction accuracy.

\section{Conclusion}
We propose a new perspective of solving few-shot segmentation with dual prototypical contrastive learning.
We design class-specific and class-agnostic contrastive loss tailored to FSS, which makes the class prototypes distinguish other classes and encourages the compact feature distribution from the same semantic class.
Our proposed DPCL achieves strong feature representation power without sub-modules required for inference.
Extensive experiments on PASCAL-$5^\text{i}$ and COCO-$20^\text{i}$ show that our method achieves state-of-the-art performance.
It demonstrates the effectiveness of integrating contrastive learning into FSS, which prevent networks to overfit the seen class and learn much more feature representations than the training class.

%%%%%%%%% REFERENCES
{\small
\bibliographystyle{ieee_fullname}
\bibliography{main}
}

\clearpage
\appendix
\onecolumn

% Define pseudocode formatting

\renewcommand{\KwSty}[1]{\textnormal{\textcolor{blue!90!black}{\ttfamily\bfseries #1}}\unskip}
\renewcommand{\ArgSty}[1]{\textnormal{\ttfamily #1}\unskip}
\SetKwComment{Comment}{\color{green!50!black}// }{}
\renewcommand{\CommentSty}[1]{\textnormal{\ttfamily\color{green!50!black}#1}\unskip}
\newcommand{\assign}{\leftarrow}
\newcommand{\var}{\texttt}
\newcommand{\FuncCall}[2]{\texttt{\bfseries #1(#2)}}
\SetKwProg{Function}{function}{}{}
\renewcommand{\ProgSty}[1]{\texttt{\bfseries #1}}

\section*{Appendix}
\vspace{5pt}

\section{Algorithms}
In the training stage, our proposed network is trained with a novel dual prototypical contrastive learning (DPCL), which consists of two terms; class-specific contrastive learning and class-agnostic contrastive learning.
\vspace{5pt}

For simplification, we explain the training and inference procedure in 1-way 1-shot scenario.
For each episode $\mathcal{E}=\{\mathcal{S},\mathcal{Q}\}$, 
we first conduct non-parametric FSS using a support set $\mathcal{S}=\{(\mathbf{I}^s, \mathbf{M}^s)\}$ and a query set $\mathcal{Q}=\{(\mathbf{I}^q, \mathbf{M}^q)\}$.
Algorithm~\ref{alg:fss} provides the pseudo-code of non-parametric FSS in a PyTorch-like style.

\begin{algorithm}[b!]
  \caption{Pseudocode of FSS}\label{alg:fss}
  \Comment{Masked Average Pooling}
  \Function{MAP(f, M)}{
  \texttt{f\_fg = bmm(f, M)}\\
  \texttt{f\_bg = bmm(f, (1-M))}\\
  \texttt{p\_fg = f\_fg.sum() / M.sum()}\\
  \texttt{p\_bg = f\_bg.sum() / (1-M).sum()}\\
  \Return{\texttt{p\_fg, p\_bg}}\\
  }
  \BlankLine
  \Comment{Non-parametric FSS function}
  \Function{\texttt{FSS(I\_s, M\_s, I\_q, M\_q, align=True):}}{
  \texttt{f\_s = enc.forward(I\_s)}\\
  \texttt{f\_q = enc.forward(I\_q)}\\
  \Comment{Generating foreground and background prototype}
  \texttt{p\_fg, p\_bg = MAP(f\_s, M\_s)}\\
  \texttt{logits\_fg = cosine\_similarity(p, p\_fg)}\\
  \texttt{logits\_bg = cosine\_similarity(p, p\_bg)}\\
  \texttt{logits = cat([logits\_ap, logits\_an])}\\
  \Comment{Estimating the query mask}
  \texttt{pred = argmax(logits)}\\
  \texttt{seg\_loss = CrossEntropyLoss(M\_q), pred)}\\
  \If{\texttt{align}}{
    \Comment{Calculating alignment loss}
    \texttt{align\_loss = FSS(I\_s, M\_s, I\_q, M\_q, align=False)}\\
    \texttt{seg\_loss += align\_loss}\\
    }
    \Return{\texttt{seg\_loss}}\
  }

\end{algorithm}

To apply the DPCL, we apply the random data augmented technique on the support and query set, followed MoCo~\cite{he2020momentum} and simCLR~\cite{chen2020simple}.
It includes random crop, random color jittering, random horizontal flip, random grayscale conversion, and random blur augmentation.
In addition, we add the momentum encoder $\mathcal{F}_{m}$ to extract consistent features from the augmented data, thus enabling overall networks to be trained stably and effectively.
\vspace{5pt}

Inspired by MoCo~\cite{he2020momentum}, the class-specific contrastive learning (CSCL) aims to make the given prototype close to its same class prototypes and far from the different class prototypes.
Using the dynamic prototype dictionary $\mathbf{D}=\{\mathbf{d}_u, \mathbf{l}_u\}_{u=1}^M$ where $M=8,192$, we set the 1 positive and 7,000 negative prototype pairs, and the detailed procedure is explained in Algorithm~\ref{alg:cscl} in a pseudo-code.
\vspace{5pt}

To extract the rich features with semantic properties, we introduce class-agnostic contrastive learning (CACL).
Given the target prototype, we set the positive and negative samples.
To construct the negative samples, we perform background random sampling and average pooling until the number of randomly sampled background features is up to 1000.
At this time, each negative key is made of 5 pixel of background features.
Note that we skip this process when the number of pixels in the down-sampled foreground is less than 5 pixels.
Algorithm~\ref{alg:cacl} provides the pseudo-code of CACL.

Leveraging the segmentation, CSCL, and CACL losses, we update our FSS networks, the feature encoder $\mathcal{F}$, the one-layer MLP $\mathcal{M}$, and momentum encoder $\mathcal{F}_{m}$.
In the inference stage, the momentum encoder $\mathcal{F}_{m}$ and the dynamic dictionary $\mathbf{D}$ are not used, and only FSS proceeds. 

\begin{algorithm}[h!]
  \caption{Pseudocode of CSCL}\label{alg:cscl}
  \Function{\texttt{CSCL(I\_s, M\_s, l\_s, D):}}{
  \Comment{Randomly augment image and mask}
  \texttt{aI\_s, aM\_s = aug(I\_s, M\_s)}\\
  \texttt{f\_s = enc.forward(I\_s)}\\
  \texttt{af\_s = enc.forward(aI\_s).detach()}\\
  \Comment{Masked average pooling}
  \texttt{p, \_ = MAP(f\_s, M\_s)~~~} \Comment{query}
  \texttt{ap, \_ = MAP(af\_s, aM\_s)~~~}   \Comment{pos}
  \texttt{an = []}\\
  \For{d\_u, l\_u in enumerate(D):}{
  \If{l\_u != l\_s:}{
  \texttt{an.append(l\_u)~~~}\Comment{neg}
  }
  }
  \texttt{logits\_ap = cosine\_similarity(p, ap)}\\
  \texttt{logits\_an = cosine\_similarity(p, an)}\\
  \texttt{logits = cat([logits\_ap, logits\_an])}\\
  \texttt{labels = zeros(logits.size(0)}\\
  \texttt{loss\_CSCL = CrossEntropyLoss(logits, labels)}\\  
  \texttt{D.append(ap)}\\
  \texttt{D.pop(ap)}\\
  \Return{\texttt{loss\_CSCL, D}}\
  }
\end{algorithm}

\begin{algorithm}[h!]
  \caption{Pseudocode of CACL}\label{alg:cacl}
  \Function{\texttt{CACL(I\_s, M\_s, I\_q, M\_q):}}{
  \Comment{Randomly augment image and mask}
  \texttt{aI\_q, aM\_q = aug(I\_q, M\_q)}\\
  \texttt{f\_s = enc.forward(I\_s)}\\
  \texttt{af\_q = enc.forward(aI\_q).detach()}\\
  \Comment{Masked average pooling}
  \texttt{p, \_ = MAP(f\_s, M\_s)~~~} \Comment{query}
  \texttt{ap, \_ = MAP(af\_q, aM\_q)~~~}   \Comment{pos}
  
  \Comment{Random average pooling for generating negative samples}
  \texttt{an = []}\\
  \For{\texttt{v in range(1000):}}{
  \texttt{an.append(RandomAP(af\_q, aM\_q))}\\
  }
  \texttt{logits\_ap = cosine\_similarity(p, ap)}\\
  \texttt{logits\_an = cosine\_similarity(p, an)}\\
  \texttt{logits = cat([logits\_ap, logits\_an])}\\
  \texttt{labels = zeros(logits.size(0)}\\
  \texttt{loss\_CACL = CrossEntropyLoss(logits, labels)}\\  
  \Return{\texttt{loss\_CACL}}\
}
\end{algorithm}

\section{More Results}
\begin{figure*}[h]
		\centering
		\includegraphics[width=0.7\textwidth]{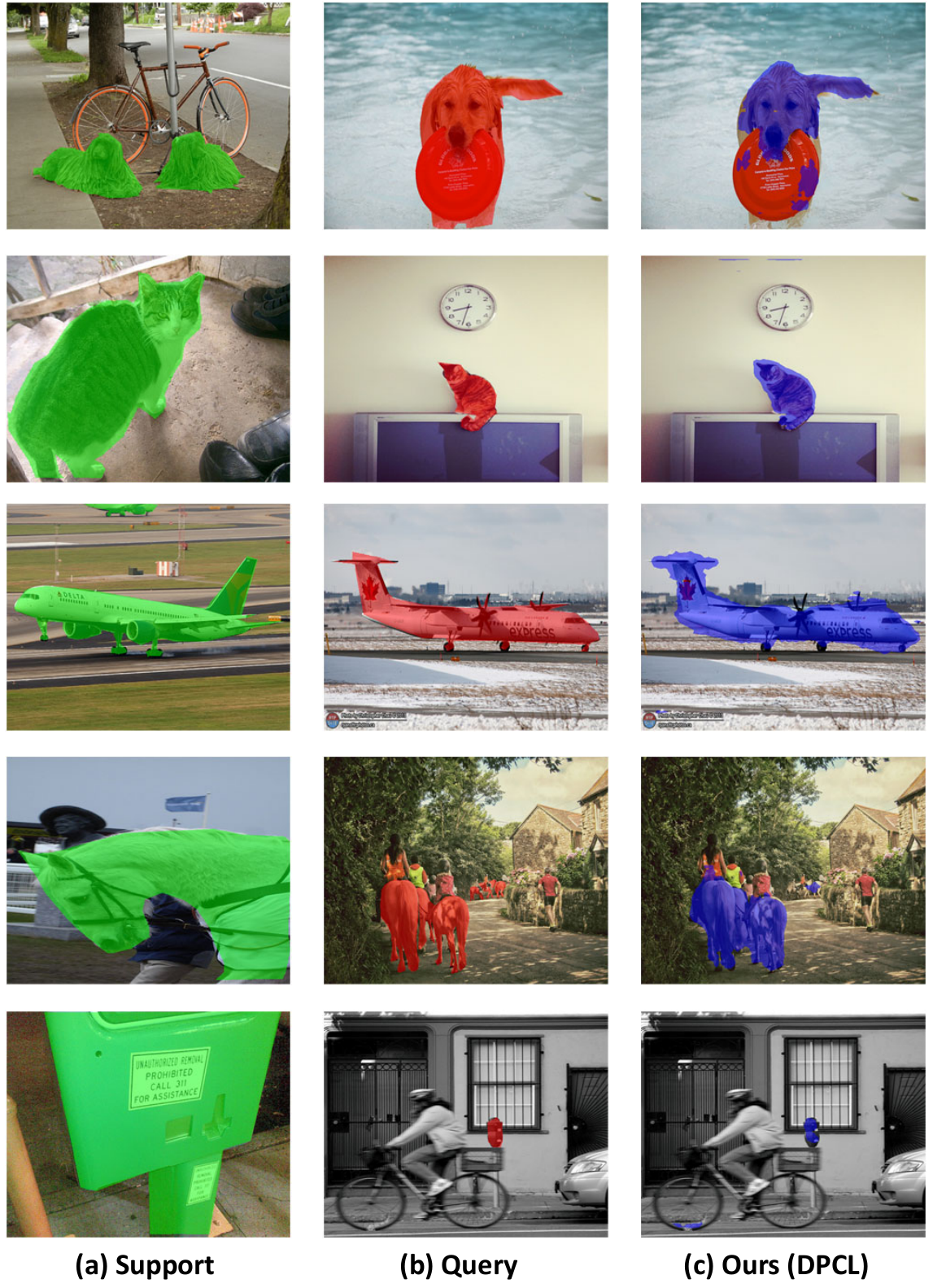}
		\caption{Qualitative results of 1-way 5-shot on COCO-$\mathbf{20}^i$.}
        \label{afig3}
\end{figure*}

\begin{figure*}[h!]
		\centering
		\includegraphics[width=0.75\textwidth]{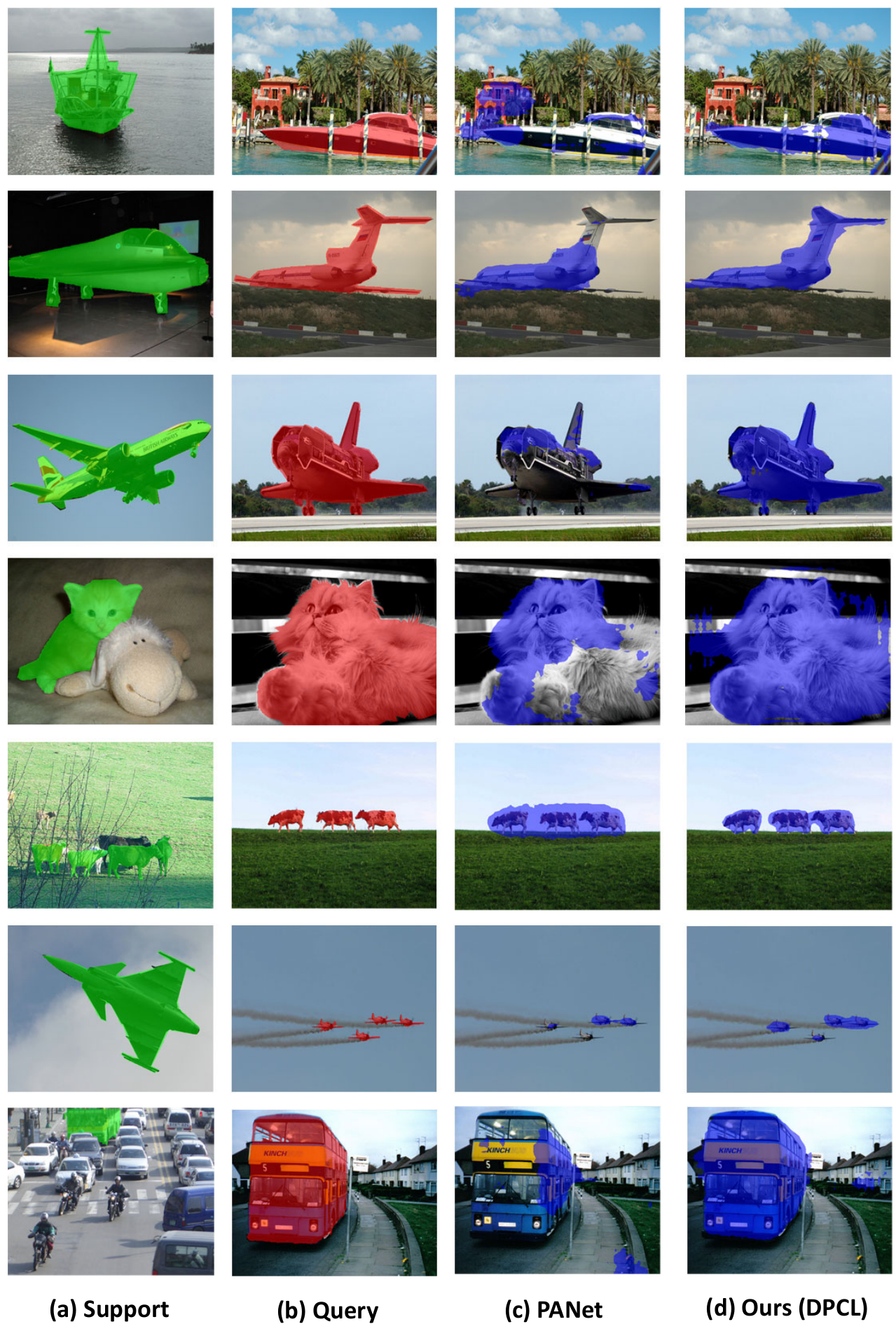}
		\caption{Qualitative results of 1-way 1-shot on PASCAL-$\mathbf{5}^i$.}
        \label{afig1}
\end{figure*}

\begin{figure*}[h]
		\centering
		\includegraphics[width=0.75\textwidth]{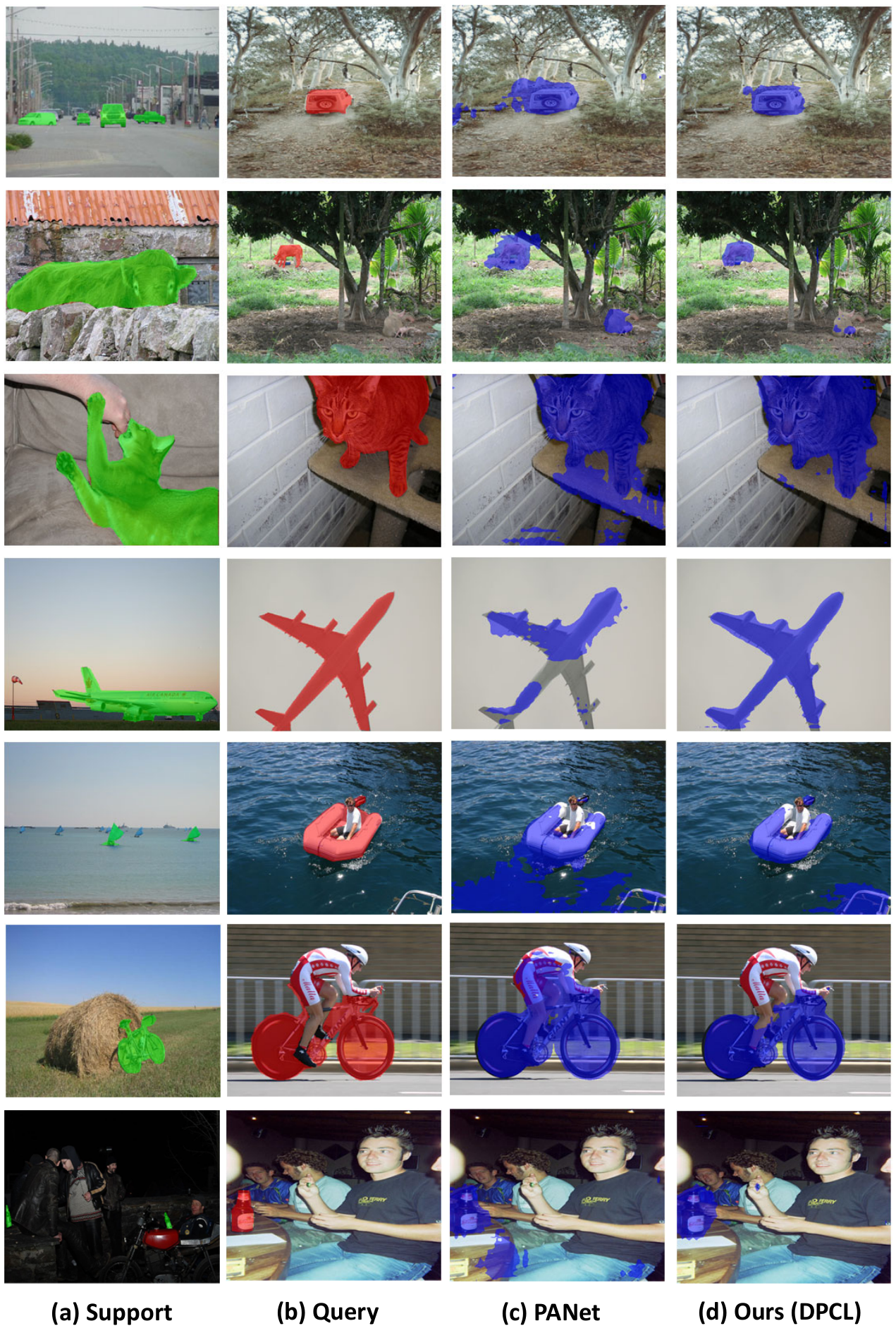}
		\caption{Qualitative results of 1-way 5-shot on PASCAL-$\mathbf{5}^i$. PANet is re-implemented with the performance from PPNet.}
        \label{afig2}
\end{figure*}

In this section, we show more qualitative results of 1-way 1-shot setting and 1-way 5-shot setting on PASCAL-$5^\text{i}$~\cite{shaban2017one} and COCO-$20^\text{i}$~\cite{wang2019panet}.
To fair evaluation, we retrained PANet~\cite{wang2019panet} based on PPNet code~\cite{liu2020part}\footnote{https://github.com/kaixin96/PANet}.
Figure~\ref{afig1} is the results of 1-way 1-shot setting and Figure~\ref{afig2} is the results of 1-way 5-shot setting on PASCAL-$5^\text{i}$~\cite{shaban2017one}.
As can be seen, our DPCL can deal with the extreme variances in shape or scale between support and query image better than PANet.
We additionally attach the 1-way 1-shot results on COCO-$20^\text{i}$~\cite{wang2019panet} in Figure~\ref{afig3}.
It shows good segmentation results on COCO-$20^\text{i}$ that is more challenging and contains diverse unseen classes, which demonstrates that our method has a superior generalization capacity.

\newpage

\newpage

\end{document}